\documentclass[10pt,twocolumn,letterpaper]{article}

\usepackage{iccv}
\usepackage{times}
\usepackage{epsfig}
\usepackage{graphicx}
\usepackage{amsmath}
\usepackage{amssymb}
\usepackage{xcolor}
\usepackage{colortbl}
\usepackage{booktabs}
\usepackage{mathtools}
\usepackage{setspace}

\usepackage[pagebackref=true,breaklinks=true,letterpaper=true,colorlinks,bookmarks=false]{hyperref}

\iccvfinalcopy 

\def\iccvPaperID{1614} 

\ificcvfinal\fi

\begin{document}

\title{InterFormer\\ Real-time Interactive Image Segmentation}

\author{
   \normalsize You Huang $^{1}$, Hao Yang $^{1}$, Ke Sun $^{1}$, Shengchuan Zhang $^{1}$\thanks{Corresponding author}, Liujuan Cao$^{1}$, Guannan Jiang$^{2}$, Rongrong Ji$^{1}$ \\
   \normalsize $^1$ Key Laboratory of Multimedia Trusted Perception and Efficient Computing, \\
   \normalsize Ministry of Education of China, Xiamen University \\
   \normalsize $^2$ Intelligent Manufacturing Department, Contemporary Amperex Technology Co. Limited (CATL)
}

\maketitle
\ificcvfinal\thispagestyle{empty}\fi

\begin{abstract}
   Interactive image segmentation enables annotators to efficiently perform pixel-level annotation for segmentation tasks. However, the existing interactive segmentation pipeline suffers from inefficient computations of interactive models because of the following two issues. 
   First, annotators' later click is based on models' feedback of annotators' former click. This serial interaction is unable to utilize model's parallelism capabilities. Second, in each interaction step, the model handles the invariant image along with the sparse variable clicks, resulting in a process that's highly repetitive and redundant. For efficient computations, we propose a method named InterFormer that follows a new pipeline to address these issues. InterFormer extracts and preprocesses the computationally time-consuming part i.e. image processing from the existing process. Specifically, InterFormer employs a large vision transformer (ViT) on high-performance devices to preprocess images in parallel, and then uses a lightweight module called interactive multi-head self attention (I-MSA) for interactive segmentation. Furthermore, the I-MSA module's deployment on low-power devices extends the practical application of interactive segmentation. The I-MSA module utilizes the preprocessed features to efficiently response to the annotator inputs in real-time. The experiments on several datasets demonstrate the effectiveness of InterFormer, which outperforms previous interactive segmentation models in terms of computational efficiency and segmentation quality, achieve real-time high-quality interactive segmentation on CPU-only devices.
   The code is available at \href{https://github.com/YouHuang67/InterFormer}{https://github.com/YouHuang67/InterFormer}.
\end{abstract}

\section{Introduction}

As fueled by massive amounts of data~\cite{ChenSun2017RevisitingUE}, deep networks achieve compelling performance in various computer vision tasks~\cite{GedasBertasius2019ClassifyingSA,HolgerCaesar2019nuScenesAM,GeertLitjens2017ASO}. The availability of accurately annotated data is essential for deep networks' success. However, the process of manual annotation is time-consuming and resource-intensive. Therefore, the developed interactive image segmentation has become an indispensable tool in annotating large-scale image datasets~\cite{RodrigoBenenson2019LargescaleIO}. Such techniques aim to achieve high-quality pixel-level annotations with limited annotator interactions, including scribbles~\cite{JunjieBai2014ErrorTolerantSB}, bounding boxes~\cite{CarstenRother2004GrabCutIF,JiajunWu2014MILCutAS}, polygons~\cite{DavidAcuna2018EfficientIA,HuanLing2019FastIO}, clicks~\cite{EirikurAgustsson2018InteractiveFI,KevisKokitsiManinis2017DeepEC,XiChen2022FocalClickTP,ZhengLinFocusCutDI,QinLiu2022PseudoClickII}, and some combinations~\cite{ShiyinZhang2020InteractiveOS}. Among them, the click-based methods have become the most prevalent due to its simplicity, and we focus on these methods in this work.

\begin{figure}[t]
   \begin{center}
      \includegraphics[width=21em,height=13em]{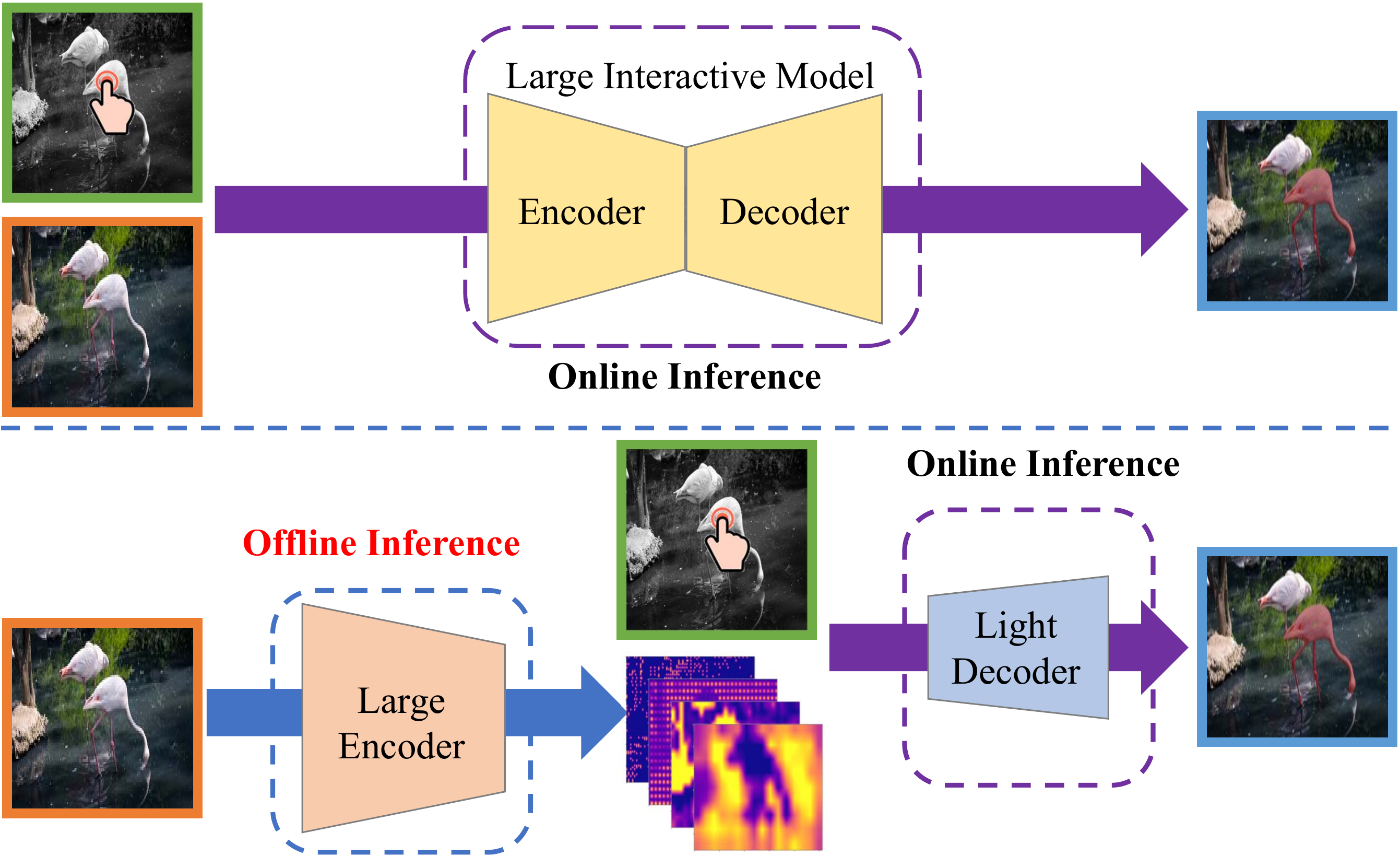}
   \end{center}
   \vspace{-1em}
      \caption{Illustration of both the existing and proposed interactive pipelines. The existing pipeline (top) provides both the image and annotator clicks to the interactive model during each interaction, while the proposed pipeline (bottom) employs a large encoder to preprocess the image and then provides the fixed encoded features and on-the-fly clicks to a light decoder during the interaction.}
   \vspace{-1em}
   \label{fig:pipeline}
   \end{figure}

Recent works on click-based interactive segmentation concentrate on various refinement modules~\cite{XiChen2022FocalClickTP,ZhengLinFocusCutDI} incorporating sophisticated engineering optimization. However, such refinement tricks still need the well-segmented results from early interactions in the interactive process. Producing such results keeps facing challenges in deploying computationally-intensive models on low-power devices. For example, it is challenging to utilize interactive segmentation models through crowdsourcing platforms~\cite{SaberMirzaeeBafti2021ACS}. Previous efforts like FocalClick~\cite{XiChen2022FocalClickTP} mitigate this issue by using lightweight models and down-sampling the inputs, but this strategy sacrifices segmentation quality as a trade-off. Hence, there is still a need for computational-friendly interactive segmentation methods on a wide range of devices.

The inefficient interactive segmentation stems from two underlying reasons. First, each annotator's click corresponds to one model inference and the next click's location depends on the former inference results. This serial interaction only processes one sample during each inference, unable to leverage parallelism capabilities of GPU. Second, throughout the annotation process on a single image, the model inputs remain notably similar, with the sparse clicks being the only variables. This leads to the extraction of near-identical features during each inference, resulting in considerable computational redundancy. Such two issues result in low computational efficiency.

In this paper, we propose a method named InterFormer to improve computational efficiency. Preceding the interactive process, InterFormer employs a large model, \eg vision transformer (ViT)~\cite{AlexeyDosovitskiy2020AnII} on high-performance devices to extract high-quality features from images to be annotated. This process is offline without the need for real-time performance. Then, InterFormer only needs a lightweight module to perform interactive segmentation on low-power devices, with such preprocessed features from a large model and clicks from annotators as inputs.

Following previous efforts~\cite{KonstantinSofiiuk2021RevivingIT} in encoding annotator clicks, we attempted to implement the lightweight module by FPN-like ConvNets~\cite{TsungYiLin2016FeaturePN} to fuse clicks with preprocessed features. However, this module fails to efficiently utilize the preprocessed features and produces unsatisfactory segmentation results (reported in Section~\ref{sec:ablationstudy}). Instead, we propose interactive multi-head self attention (I-MSA), an efficient interactive segmentation module, inspired by the recent success of ViT's variants~\cite{liu2021swin,WenhaiWang2021PyramidVT,YuHuanWu2021P2TPP}. I-MSA has extremely low computational complexity and is optimized for high utilization of the preprocessed features from the large models. Furthermore, we adopt the Zoom-In strategy~\cite{KonstantinSofiiuk2020fBRSRB} and slightly modify the deeper blocks of I-MSA to focus on the valid regions around the potential objects of images. Such modified blocks lead to significantly faster inference of I-MSA with slight drop in performance. 

\begin{figure}[t]
\begin{center}
   \includegraphics[width=0.8\linewidth]{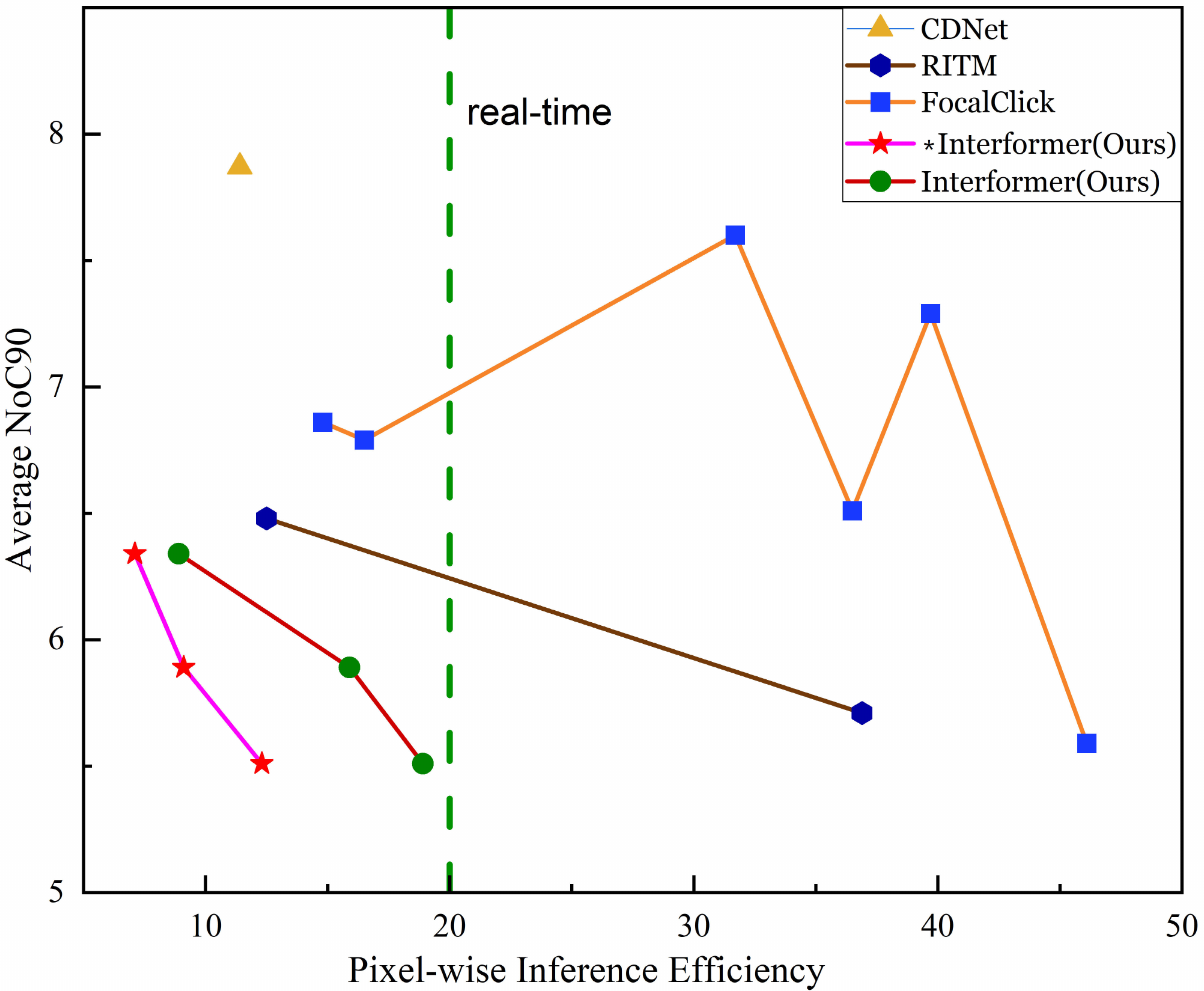}
\end{center}
   \caption{Experimental results for interactive segmentation on SBD dataset~\cite{BharathHariharan2011SemanticCF}. The average NoC90 indicates the average number of clicks required to achieve IoU of 0.9, and the PIE (pixel-wise inference efficiency) measures a model's efficiency in pixel-wise segmentation across potentially large or small objects. Our proposed InterFormer ($*$ indicates the measure of PIE without considering the preprocessing) showcases faster inference with better segmentation results, than the recently proposed methods.}
\label{fig:speed}
\vspace{-1em}
\end{figure}

As illustrated in Figure~\ref{fig:speed}, the proposed InterFormer outperforms the previous interactive segmentation models and achieves state-of-the-art performance at the similar computational cost. The measure of computation speeds takes into account the process of ViT extracting features (on the same device) for fair comparison. Moreover, InterFormer achieves real-time high-quality interactive segmentation on cpu-only devices, based on the offline preprocessed features. We extensively conduct experiments on GrabCut~\cite{CarstenRother2004GrabCutIF}, Berkeley~\cite{KevinMcGuinness2010ACE}, SBD~\cite{BharathHariharan2011SemanticCF}, and DAVIS~\cite{FedericoPerazzi2016ABD} datasets to substantiate the effectiveness of InterFormer.

We summarize our contributions as follows:
\begin{itemize}
   \item We introduce a method called InterFormer, which follows a new pipeline splitting the interactive process into two stages to take advantage of well-developed large-scale models and accelerate the interaction.
   \item We propose an interactive attention module named I-MSA utilizing the prepropressed features to achieve high-quality real-time interactive segmentation on cpu-only devices.
   \item InterFormer significantly outperforms the previous methods in terms of computational efficiency and segmentation quality.
\end{itemize}

\section{Related Work}\label{sec:relatedwork}

\subsection{Interactive Segmentation}
Before the advent of deep networks, graph-based methods~\cite{YuriBoykov2001InteractiveGC,LeoGrady2006RandomWF,CarstenRother2004GrabCutIF} were prevalent in interactive image segmentation research. These methods formulate the image as a graph and use optimization methods to solve the graph cut problem given annotator inputs. However, the low-level features used by such methods are limited to vanilla cases of image segmentation. DIOS~\cite{NingXu2016DeepIO} was the first to introduce deep networks into interactive segmentation and proposed a strategy to transform clicks into distance maps concatenated to the image. DIOS also formulated the training/test pipeline for click-based methods. Several methods have been proposed to enhance click-based interactive segmentation, including DEXTR~\cite{KevisKokitsiManinis2017DeepEC}, FCA-Net~\cite{ZhengLin2020InteractiveIS}, BRS~\cite{JangWonDong2019InteractiveIS} and f-BRS~\cite{KonstantinSofiiuk2020fBRSRB}. These methods focus on different aspects of interactive segmentation to improve efficiency, \eg the four extreme points around the object~\cite{KevisKokitsiManinis2017DeepEC}, the first click~\cite{ZhengLin2020InteractiveIS}, and the inference optimization~\cite{JangWonDong2019InteractiveIS,KonstantinSofiiuk2020fBRSRB}. More recently, RITM~\cite{KonstantinSofiiuk2021RevivingIT} was proposed, which incorporates the previous segmentation result into the model inputs. Other recent click-based methods~\cite{XiChen2022FocalClickTP,ZhengLinFocusCutDI} focus on local refinement and decompose the previous pipeline into coarse segmentation and refinement implemented by lightweight models. Besides, PseudoClick~\cite{QinLiu2022PseudoClickII} simulates annotator clicks using an additional module and corresponding loss. In this paper, we propose a new pipeline that differs from existing methods. Our pipeline preprocesses the image offline using large models and performs interactive segmentation using lightweight models.

\begin{figure*}[ht]
   \begin{center}
      \includegraphics[width=1.0\linewidth]{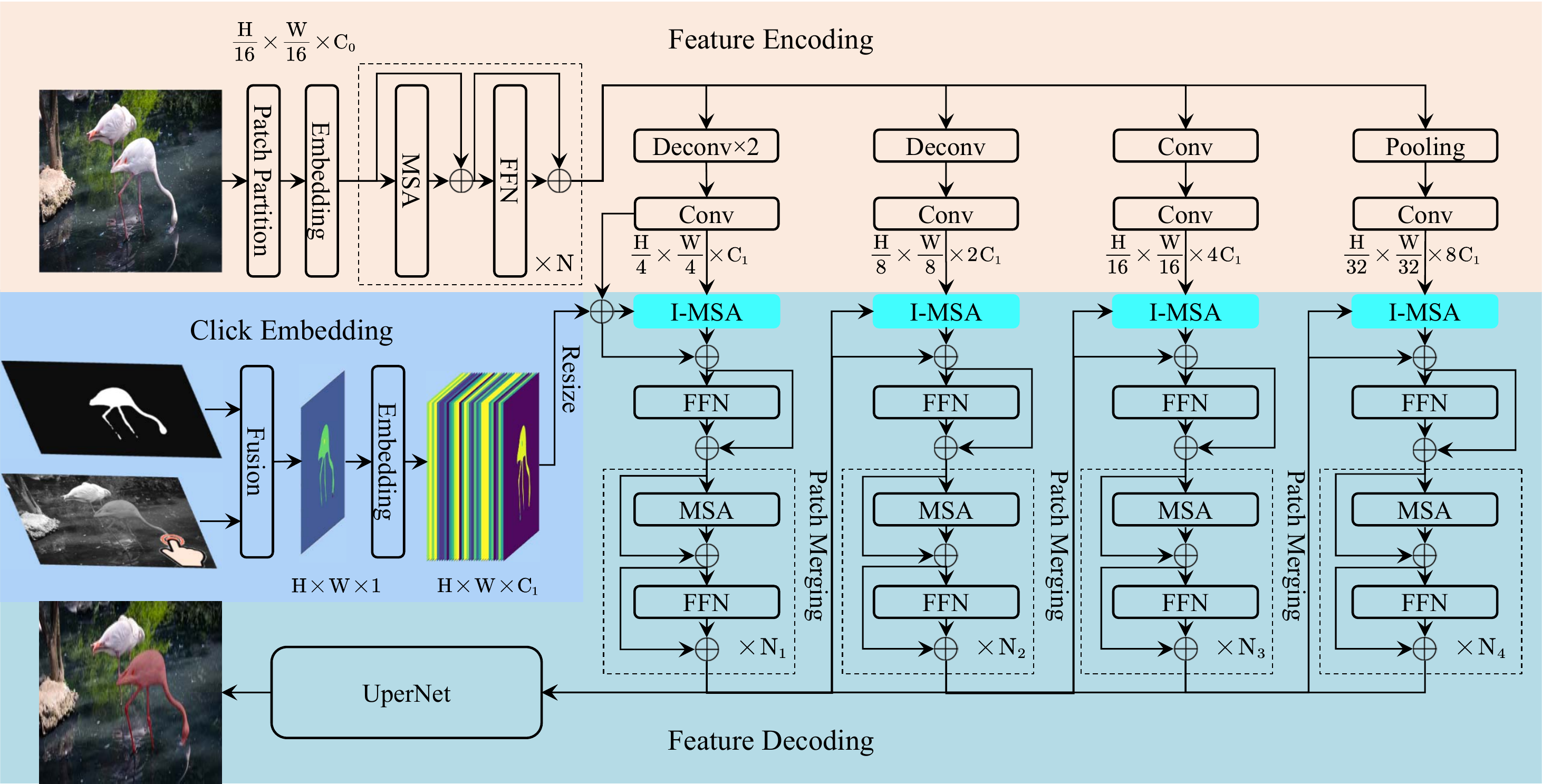}
   \end{center}
      \caption{The overview of our proposed InterFormer. The process separates feature extraction from interaction. First, the feature encoding module extracts multi-scale features from the input image. Next, the click embedding module generates a click map and fuses it with the image features. Finally, the feature decoding module integrates the multi-scale features and decodes them into final segmentation results.}
   \label{fig:architecture}
   \end{figure*}

\subsection{Vision Transformer}
The Transformer architecture was originally proposed for machine translation~\cite{AshishVaswani2017AttentionIA}, and is designed to model global token-to-token dependencies. This modeling approach was then adapted for computer vision tasks by the Vision Transformer (ViT)~\cite{AlexeyDosovitskiy2020AnII} which links patches distributed around the entire image. This inspired a series of ViT-based works~\cite{HugoTouvron2020TrainingDI,LiYuan2021TokenstoTokenVT,WenhaiWang2021PyramidVT,HaoqiFan2021MultiscaleVT,liu2021swin,YuHuanWu2021P2TPP,hu2023you,hu2021istr} that have successfully tackled various computer vision tasks. Subsequently, ViTs were further developed for image segmentation, \eg SETR~\cite{SixiaoZheng2021RethinkingSS}, Segmenter~\cite{RobinStrudel2021SegmenterTF} and SegFormer~\cite{EnzeXie2021SegFormerSA}. More recently, FocalClick~\cite{XiChen2022FocalClickTP} employed SegFormer for interactive segmentation, and SimpleClick~\cite{QinLiu2022SimpleClickII} introduced the MAE-pretrained ViT~\cite{KaimingHe2021MaskedAA} into interactive segmentation. In this paper, we followed the approach of SimpleClick~\cite{QinLiu2022SimpleClickII} and use a large plain ViT as our encoder for feature encoding. This was performed offline without the need for real-time performance and resulted in high-quality preprocessed features.

\section{Method}\label{sec:method}

We propose InterFormer that follows a computationally efficient pipeline to address the inefficiencies of the existing pipeline. Section~\ref{sec:pipeline} introduces the new pipeline. Section~\ref{sec:imsa} describes the proposed I-MSA modules to implement this pipeline. Section~\ref{sec:zoomin} provides a zoom-in strategy of InterFormer. Section~\ref{sec:training} concludes a simplified training setting. 

\subsection{Computationally Efficient Pipeline}\label{sec:pipeline}
The proposed pipeline decomposes the interaction based on large models into three distinct processes: feature encoding, click embedding, and feature decoding (as illustrated in Figure~\ref{fig:architecture}). Feature encoding facilitates offline preprocessing of images by large models to extract reusable features, thereby obviating the need for real-time performance. During the interaction, this pipeline transforms the annotator clicks into embeddings, and then provides the decoder with the preprocessed features and embeddings to produce the segmentation results.

\textbf{Feature encoding.}
Inspired by the work of SimpleClick~\cite{QinLiu2022SimpleClickII}, we employ large models for feature encoding, such as the widely used MAE-pretrained Vision Transformer (ViT)~\cite{KaimingHe2021MaskedAA}. Besides, following the approach in ViTDet~\cite{YanghaoLiExploringPV}, we use a simple FPN~\cite{TsungYiLin2016FeaturePN} to produce multi-scale features from the single-scale patch embeddings in the plain ViT's last block. For instance, ViT-Base (ViT-B)~\cite{KaimingHe2021MaskedAA} patchifies the input image of size $H\times W$ into a sequence of $16\times16$ patches, which are then projected into $C_0$-dimensional vectors. The ViT blocks perform multi-head self attention on these vectors with fixed length, producing $\frac{H}{16}\times\frac{W}{16}$ $C_0$-dimensional vectors. We use a simple FPN to convert these vectors into feature maps $\boldsymbol{F}_i, 1\le i \le 4$, with channels of $2^{i-1} C_1, 1 \le i \le 4$, and sizes of $\frac{H}{4}\times\frac{H}{4}$, $\frac{H}{8}\times\frac{H}{8}$, $\frac{H}{16}\times\frac{H}{16}$, and $\frac{H}{32}\times\frac{H}{32}$, respectively, through multiple convolution and pooling layers. This module simultaneously enables multi-scale information extraction and cuts down the computational expense of subsequent decoding.

\textbf{Click Embedding.}
During interactions, annotators provide a single click in the erroneous region of the model prediction, corresponding to false negative or false positive regions for positive and negative clicks, respectively. Previous methods~\cite{KonstantinSofiiuk2021RevivingIT,XiChen2022FocalClickTP} encode such clicks by adding two maps to the image channels, consisting of disks around the clicks. This strategy slightly modifies the successive click maps. The RITM pipeline~\cite{KonstantinSofiiuk2021RevivingIT} further improves the segmentation by augmenting the image channels with previous segmentation results. We adopt these strategies and fuse the click maps with previous results instead of adding more channels. 

We maintain a mask $\boldsymbol{M}_{\text{ref}}$ of size $H\times W$ and update it at each interaction. Each pixel in $\boldsymbol{M}_{\text{ref}}$ is categorized as either $[D_{\text{fg}}]$, $[P_{\text{fg}}]$, $[U]$, $[P_{\text{bg}}]$, or $[D_{\text{bg}}]$ based on its foreground/background confidence, where ``$P$'' stands for ``possible'', ``$D$'' signifies ``definite'', ``bg'' refers to the background, ``fg'' refers to the foreground, and ``U'' represents ``Unknown''. 
Initially, all pixels in $\boldsymbol{M}_{\text{ref}}$ are set to $[U]$. During the interaction process, each new click introduces a disk, with a radius of five pixels, to $\boldsymbol{M}_{\text{ref}}$. The pixels within this disk are labeled as either $[D_{\text{fg}}]$ or $[D_{\text{bg}}]$, determined by whether the click is positive or negative. After the click is incorporated, the model predicts the segmentation results, assigning the labels $[P_{\text{fg}}]$ or $[P_{\text{bg}}]$ to the pixels identified as foreground or background. The labels within the click disks remain unchanged. The pixels outside these disks are assigned one of the labels $[P_{\text{fg}}]$, $[P_{\text{bg}}]$, or $[U]$, based on the current state of $\boldsymbol{M}_{\text{ref}}$ and the segmentation prediction. The assignment follows the rules: $[U] + [P_{\text{fg}}] \rightarrow [P_{\text{fg}}]$, $[U] + [P_{\text{bg}}] \rightarrow [P_{\text{bg}}]$, and $[P_{\text{fg}}] + [P_{\text{bg}}] \rightarrow [U]$.

During the model inference, each label category in $M_{\text{ref}}$ corresponds to a specific learnable $C_1$-dimensional vector. This mapping transforms $M_{\text{ref}}$ into a click embedding $\boldsymbol{E}_c \in \mathcal{R}^{H\times W\times C_1}$. To create the click-involved feature $\boldsymbol{F}_c$, we then resize $\boldsymbol{E}_c$ and combine it with the feature $\boldsymbol{F}_1$ derived from the FPN module: $\boldsymbol{F}_c = \text{resize}(\boldsymbol{E}_c) + \boldsymbol{F}_1$.

\textbf{Feature Decoding.}
We further fuse the encoded features $\boldsymbol{F}_i, 1 \le i \le 4$ with the click-involved feature $\boldsymbol{F}_c$ using the proposed I-MSA module (introduced in Section~\ref{sec:imsa}). The fusion is then decoded to generate the final segmentation results using the widely used UperNet~\cite{TeteXiao2018UnifiedPP}.

\subsection{Interactive Multi-head Self Attention}\label{sec:imsa}
Previous works~\cite{KonstantinSofiiuk2021RevivingIT,XiChen2022FocalClickTP,ZhengLinFocusCutDI,QinLiu2022PseudoClickII} have used a vanilla approach to process simply encoded click maps by extending them as additional channels to the image. However, this approach requires a deep stack of convolution layers~\cite{KonstantinSofiiuk2021RevivingIT,ZhengLinFocusCutDI,QinLiu2022PseudoClickII} or self attention blocks~\cite{XiChen2022FocalClickTP} to effectively fuse the sparse click information with the image features, resulting in high computational complexity. In our pipeline, this would lead to degenerated results due to the shallow architecture of the decoder (reported in Section~\ref{sec:ablationstudy}). Instead, we propose I-MSA to efficiently incorporate the click information.

\textbf{Interactive Self Attention.}
With the emergence of transformers, there has been increased attention on the self attention mechanism, which leverages patch embeddings' similarity to single out task-critical patches~\cite{AlexeyDosovitskiy2020AnII}. Interactive segmentation naturally fits into this mechanism, as annotator clicks provide the regions of interest, and the model can use these features to identify the components of the object of interest based on their similarity. This inspired us to reformulate the self attention as an interactive one. Specifically, we reposition the click-involved feature $\boldsymbol{F}_c$ and the preprocessed features $\boldsymbol{F}_i, 1 \le i \le 4$, as the query $\boldsymbol{Q}$ and key $\boldsymbol{K}$ in self attention, with the value $\boldsymbol{V}$ sharing the same features as the key. For convenience, we reformulate the traditional $(\boldsymbol{Q}, \boldsymbol{K}, \boldsymbol{V})$~\cite{AlexeyDosovitskiy2020AnII} as a function
\begin{equation}
   (\boldsymbol{Q}, \boldsymbol{K}, \boldsymbol{V})(\boldsymbol{A}, \boldsymbol{B}) = (\boldsymbol{A}\boldsymbol{W}^q, \boldsymbol{B}\boldsymbol{W}^k, \boldsymbol{B}\boldsymbol{W}^v),
\end{equation}
given the inputs $\boldsymbol{A}$ and $\boldsymbol{B}$. Then, we define the regular multi-head self attention module as $\text{MSA}(\boldsymbol{A}) = (\boldsymbol{Q}, \boldsymbol{K}, \boldsymbol{V})(\boldsymbol{A}, \boldsymbol{A})$, and the proposed I-MSA module as $\text{I-MSA}(\boldsymbol{A}, \boldsymbol{B}) = (\boldsymbol{Q}, \boldsymbol{K}, \boldsymbol{V})(\boldsymbol{A}, \boldsymbol{B})$. The tuple $(\boldsymbol{Q}, \boldsymbol{K}, \boldsymbol{V})$ is utilized for regular attention computation via a softmax operator. For brevity, we omit detailing this operation, assuming that attention computation is inherent to the modules presented.

As outlined in Figure~\ref{fig:architecture}, we adopt the hierarchical architecture of recent ViT variants~\cite{liu2021swin,WenhaiWang2021PyramidVT} to construct hierarchical representation by starting from low-level features $\boldsymbol{F}_1$. We perform interactive attention on $\boldsymbol{F}_1$ and $\boldsymbol{F}_c$ to get $\boldsymbol{H}_1^1 = \text{I-MSA}(\boldsymbol{F}_c, \boldsymbol{F}_1)$. Then, we perform self attention on $\boldsymbol{H}_1^1$ iteratively. In other words, we compute $\boldsymbol{H}_1^i = \text{MSA}(\boldsymbol{H}_1^{i-1})$ for each $2 \le i \le N_1 + 1$, where $N_1$ is the depth of the first stage.  Subsequently, we employ interactive attention on $\boldsymbol{F}_2, \boldsymbol{H}_1^{N_1 + 1}$ to start the second stage instead of reusing the vanilla click-involved feature $\boldsymbol{F}_c$. Besides, before initiating the second stage, we adopt the patch merging operation~\cite{liu2021swin} used in the hierarchical Transformers on $\boldsymbol{H}_1^{N_1+1}$. The subsequent two stages corresponding to $\boldsymbol{F}_3$ and $\boldsymbol{F}_4$ respectively follow the same process. Finally, we obtain the features $\boldsymbol{H}_i^{N_i + 1}, 1 \le i \le 4$, which are then fed into the decoder UperNet.

\textbf{Pooling-based self attention.}
We have further improved the efficiency of our interactive self attention by adopting P2T's pooling strategy~\cite{YuHuanWu2021P2TPP}. This technique helps to alleviate the quadratic computational complexity of self attention. Specifically, P2T uses a series of average pooling layers with different pooling ratios to preprocess the input $\boldsymbol{B}$ of $(\boldsymbol{Q}, \boldsymbol{K}, \boldsymbol{V})(\boldsymbol{A}, \boldsymbol{B})$, \ie 
\begin{equation}
   \begin{split}
      \boldsymbol{P}_1 = \text{AvgPool}_1(\boldsymbol{B}), \\
      \boldsymbol{P}_2 = \text{AvgPool}_2(\boldsymbol{B}), \\
      \ldots, \\
      \boldsymbol{P}_n = \text{AvgPool}_n(\boldsymbol{B}). \\
   \end{split}
\end{equation}
The resulting pyramid features $\boldsymbol{P}_i, 1 \le i \le n$ are then processed using depth-wise convolution, flattened, and concatenated to produce a shorter sequence of features. Finally, these pooled features are fed into $(\boldsymbol{Q}, \boldsymbol{K}, \boldsymbol{V})(\boldsymbol{A}, \cdot)$, with $\boldsymbol{A}$ remaining unchanged.

\subsection{Zoom-in Strategy}\label{sec:zoomin}
The zoom-in strategy proposed in~\cite{KonstantinSofiiuk2020fBRSRB} involves cropping and resizing the area around the potential object in the original image to a sufficient size for segmentation models. However, the proposed pipeline is incompatible with the zoom-in strategy~\cite{KonstantinSofiiuk2020fBRSRB} since feature extraction is completed in the offline preprocessing stage, and it is impossible to preprocess the crops around the object as required. Therefore, we have made slight modifications to the I-MSA module to incorporate this strategy in feature decoding.

In details, the zoom-in strategy is a technique for selecting a region of interest (RoI) in image segmentation tasks, which involves identifying the foreground using a bounding box and annotator clicks. To ensure adequate context, the RoI is enlarged by a factor of 1.4. The resulting RoI coordinates are projected onto the coordinate system of the feature map $\boldsymbol{F}_i$ through dividing the coordinates by the stride of the feature (\eg 4, 8, 16, or 32). The relevant features are then cropped and passed through I-MSA module. Due to discontinuity of features, we slightly enlarge the RoI to ensure divisibility by the maximum stride (\ie 32), thereby facilitating feature cropping without interpolation strategies such as RoIAlign~\cite{KaimingHe2017MaskR}. Besides, we only feed the cropped features into the deeper blocks of each stage and inject the transformed cropped features into the full features.

\subsection{Training}\label{sec:training}

\textbf{Click Simulation.}
We simplify the click simulation approach proposed by RITM~\cite{KonstantinSofiiuk2021RevivingIT} to train our InterFormer model. The original approach involves randomly sampling clicks inside or outside the ground truth masks to simulate the interaction process, followed by iterative generation of clicks based on the segmented results. However, we found that eliminating the initial random click sampling and only performing iterative simulation was sufficient for our needs. To balance the computational cost with the need for a sufficient number of simulations, we use an exponentially decaying probability to sample the number of simulations. Our approach achieve comparable computational speeds to the original simulation process, without requiring specialized design of the simulation strategy.

\textbf{Training supervision.}
We adopt the normalized focal loss (NFL) proposed in RITM~\cite{KonstantinSofiiuk2021RevivingIT} that is proved to have better convergence of training interactive segmentation models, and we further empirically demonstrate NFL's effectiveness on our models.

\section{Experiments}\label{sec:experiments}
In Section~\ref{sec:experimentalsetting}, we outline the basic settings and training/test details of the proposed InterFormer. In Section~\ref{sec:mainresult}, we compare the performance of InterFormer with other existing methods on various benchmark datasets including GrabCut~\cite{CarstenRother2004GrabCutIF}, Berkeley~\cite{KevinMcGuinness2010ACE}, SBD~\cite{BharathHariharan2011SemanticCF}, and DAVIS~\cite{FedericoPerazzi2016ABD} datasets. We report the results of our ablation studies in Section~\ref{sec:ablationstudy} to analyze the impact of different components of InterFormer. Finally, in Section~\ref{sec:qualitativeresult}, we present qualitative results of InterFormer, showcasing its effectiveness for interactive segmentation. 

\subsection{Experimental Setting}\label{sec:experimentalsetting}

\begin{table}[htbp]
   \centering
   \begin{tabular}{lcc}
      \toprule
      Model & InterFormer-Light & InterFormer-Tiny \\
      \midrule
      Backbone & ViT-Base & ViT-Large \\
      $N_1, N_2, N_3, N_4$ & 0, 0, 1, 0 & 1, 1, 5, 2 \\
      \bottomrule
      \end{tabular}%
   \vspace{1em}
   \caption{Configurations of InterFormers.}
   \label{tab:configs}%
 \end{table}%

 \begin{table*}[ht]
   \begin{center}
   \scalebox{0.8}{
   \begin{tabular}{ll|c|c|c|c|c|c|c|c}
   \toprule[1pt]
   \multicolumn{3}{l|}{} &  &  & GrabCut~\cite{CarstenRother2004GrabCutIF} & Berkeley~\cite{KevinMcGuinness2010ACE} & \multicolumn{2}{c|}{SBD~\cite{BharathHariharan2011SemanticCF}} & DAVIS~\cite{FedericoPerazzi2016ABD} \\
   \cline{6-10}
   \multicolumn{2}{l|}{Method } & Train Data & PIE (1e-7) & SPC (Size) & NoC~90 & NoC~90 & NoC~85 & NoC~90 & NoC~90 \\
   \hline
   \multicolumn{2}{l|}{Graph cut~\cite{YuriBoykov2001InteractiveGC}} & /  & -- & -- & 10.00 & 14.22 & 13.6 & 15.96 & 17.41 \\
   \multicolumn{2}{l|}{Geodesic matting~\cite{VarunGulshan2010GeodesicSC}} & / & -- & -- & 14.57 & 15.96 & 15.36 & 17.60 & 19.50 \\
   \multicolumn{2}{l|}{Random walker~\cite{LeoGrady2006RandomWF}}  & / & -- & -- & 13.77 & 14.02 & 12.22 & 15.04 & 18.31 \\
   \multicolumn{2}{l|}{Euclidean star convexity~\cite{VarunGulshan2010GeodesicSC}}  & /  & -- & -- & 9.20 & 12.11 & 12.21 & 14.86 & 17.70 \\
   \multicolumn{2}{l|}{Geodesic star convexity~\cite{VarunGulshan2010GeodesicSC}}  & / & -- & -- & 9.12 & 12.57 & 12.69 & 15.31 & 17.52 \\
   \hline
   \multicolumn{2}{l|}{DOS w/o GC~\cite{NingXu2016DeepIO}}  & \scalebox{0.75}{ Augmented VOC~\cite{MarkEveringham2010ThePV,BharathHariharan2011SemanticCF}} & -- & -- & 12.59 & -- & 14.30 & 16.79 & 17.11 \\
   \multicolumn{2}{l|}{DOS with GC~\cite{NingXu2016DeepIO}}  & \scalebox{0.75}{ Augmented VOC~\cite{MarkEveringham2010ThePV,BharathHariharan2011SemanticCF}} & -- & -- & 6.08 & -- & 9.22 & 12.80 & 12.58 \\
   \multicolumn{2}{l|}{RIS-Net~\cite{JunHaoLiew2021DeepIT}}  & \scalebox{0.75}{ Augmented VOC~\cite{MarkEveringham2010ThePV,BharathHariharan2011SemanticCF}}& -- & -- & 5.00 & -- & 6.03 & -- & -- \\
   \multicolumn{2}{l|}{CM guidance~\cite{MajumderSoumajit2019ContentAwareMG}}  & \scalebox{0.75}{ Augmented VOC~\cite{MarkEveringham2010ThePV,BharathHariharan2011SemanticCF}}& -- & -- & 3.58 & 5.60 & -- & -- & --\\
   \multicolumn{2}{l|}{FCANet~(SIS)~\cite{ZhengLin2020InteractiveIS}}  & \scalebox{0.75}{ Augmented VOC~\cite{MarkEveringham2010ThePV,BharathHariharan2011SemanticCF}} & -- & -- & 2.14 & 4.19 & --  & --  & 7.90 \\
   \multicolumn{2}{l|}{Latent diversity~\cite{ZhuwenLi2018InteractiveIS}}  & \scalebox{0.75}{SBD~\cite{BharathHariharan2011SemanticCF}} & -- & -- & 4.79 & -- & 7.41 & 10.78 & 9.57 \\
   \multicolumn{2}{l|}{BRS~\cite{JangWonDong2019InteractiveIS}}  & \scalebox{0.75}{SBD~\cite{BharathHariharan2011SemanticCF}} & -- & -- & 3.60 & 5.08 & 6.59 & 9.78 & 8.24 \\
   \multicolumn{2}{l|}{f-BRS-B-resnet50~\cite{KonstantinSofiiuk2020fBRSRB}}  & \scalebox{0.75}{SBD~\cite{BharathHariharan2011SemanticCF}} & -- & -- & 2.98 & {4.34} & 5.06 & 8.08 & 7.81 \\
   \multicolumn{2}{l|}{CDNet-resnet50~\cite{XiChen2021ConditionalDF}}  & \scalebox{0.75}{SBD~\cite{BharathHariharan2011SemanticCF}} & 11.4 & 0.30 (512) & 2.64 & 3.69 & 4.37 &  7.87 & 6.66 \\
   \multicolumn{2}{l|}{ RITM-HRNet18~\cite{KonstantinSofiiuk2021RevivingIT}}  & \scalebox{0.75}{SBD~\cite{BharathHariharan2011SemanticCF}} & -- & -- & 2.04 & 3.22 & 3.39 & 5.43 & 6.71 \\
   \multicolumn{2}{l|}{ FocalClick-HRNet18s-S2~\cite{XiChen2022FocalClickTP} } &  \scalebox{0.75}{SBD~\cite{BharathHariharan2011SemanticCF}} & 16.5 & 0.11 (256) & 2.06 & 3.14 & 4.30 & 6.52 & 6.48  \\
   \multicolumn{2}{l|}{ FocalClick-SegFormerB0-S2~\cite{XiChen2022FocalClickTP}}  & \scalebox{0.75}{SBD~\cite{BharathHariharan2011SemanticCF}} & 14.8 & 0.10 (256) & 1.90 & 3.14 & 4.34 & 6.51 & 7.06 \\
   \multicolumn{2}{l|}{ FocusCut-ResNet-50~\cite{ZhengLinFocusCutDI}}  & \scalebox{0.75}{SBD~\cite{BharathHariharan2011SemanticCF}}  & -- & -- & 1.78 & 3.44 & 3.62 & 5.66 & 6.38 \\
   \multicolumn{2}{l|}{ FocusCut-ResNet-101~\cite{ZhengLinFocusCutDI}}  & \scalebox{0.75}{SBD~\cite{BharathHariharan2011SemanticCF}}  & -- & -- & 1.64 & 3.01 & 3.40 & 5.31 & 6.22 \\
   \multicolumn{2}{l|}{ PseudoClick-HRNet18~\cite{QinLiu2022PseudoClickII}}  & \scalebox{0.75}{SBD~\cite{BharathHariharan2011SemanticCF}}  & -- & -- & 2.04 & 3.23 & -- & 5.40 & 6.57 \\
   \multicolumn{2}{l|}{ PseudoClick-HRNet32~\cite{QinLiu2022PseudoClickII}}  & \scalebox{0.75}{SBD~\cite{BharathHariharan2011SemanticCF}}  & -- & -- & 1.84 & 2.98 & -- & 5.61 & 6.16 \\
   \hline
   \multicolumn{2}{l|}{ 99\%AccuracyNet~\cite{MarcoForte2020GettingT9}}  & \scalebox{0.75}{Synthetic~\cite{DengxinDai2014TheSO,BharathHariharan2011SemanticCF,TsungYiLin2014MicrosoftCC,NingXu2017DeepIM}}   & -- & -- & 1.80 & 3.04 & 3.90 & -- & -- \\
   \multicolumn{2}{l|}{f-BRS-B-HRNet32~\cite{KonstantinSofiiuk2020fBRSRB}}  & \scalebox{0.75}{COCO~\cite{TsungYiLin2014MicrosoftCC}+LVIS~\cite{AgrimGupta2019LVISAD}}  & -- & -- & 1.69 & 2.44 & 4.37 & 7.26 & 6.50 \\
   \multicolumn{2}{l|}{ RITM-HRNet18s~\cite{KonstantinSofiiuk2021RevivingIT}}  & \scalebox{0.75}{COCO~\cite{TsungYiLin2014MicrosoftCC}+LVIS~\cite{AgrimGupta2019LVISAD}} & 12.5 & 0.20 (400) & 1.68 & 2.60 & 4.04 & 6.48 & 5.98 \\
   \multicolumn{2}{l|}{ RITM-HRNet32~\cite{KonstantinSofiiuk2021RevivingIT}}  & \scalebox{0.75}{COCO~\cite{TsungYiLin2014MicrosoftCC}+LVIS~\cite{AgrimGupta2019LVISAD}} & 36.9 & 0.59 (400) & 1.56 & 2.10 & 3.59 & 5.71 & 5.34 \\
   \multicolumn{2}{l|}{ EdgeFlow-HRNet18~\cite{YuyingHao2021EdgeFlowAP}}  & \scalebox{0.75}{COCO~\cite{TsungYiLin2014MicrosoftCC}+LVIS~\cite{AgrimGupta2019LVISAD}} & -- & -- & 1.72 & 2.40 & -- & -- & 5.77 \\
   \multicolumn{2}{l|}{ FocalClick-HRNet18s-S1~\cite{XiChen2022FocalClickTP}} &  \scalebox{0.75}{COCO~\cite{TsungYiLin2014MicrosoftCC}+LVIS~\cite{AgrimGupta2019LVISAD}} & 39.7 & 0.07 (128) & 1.82 & 2.89 & 4.74 & 7.29 & 6.56 \\
   \multicolumn{2}{l|}{ FocalClick-HRNet18s-S2~\cite{XiChen2022FocalClickTP} } &  \scalebox{0.75}{COCO~\cite{TsungYiLin2014MicrosoftCC}+LVIS~\cite{AgrimGupta2019LVISAD}} & 16.5 & 0.11 (256) & 1.62 & 2.66 & 4.43 &  6.79 & 5.25 \\
   \multicolumn{2}{l|}{ FocalClick-HRNet32-S2~\cite{XiChen2022FocalClickTP} } &  \scalebox{0.75}{COCO~\cite{TsungYiLin2014MicrosoftCC}+LVIS~\cite{AgrimGupta2019LVISAD}} & 36.5 & 0.24 (256) & 1.80  & 2.36 & 4.24 &  6.51 & 5.39 \\
   \multicolumn{2}{l|}{ FocalClick-SegFormerB0-S1~\cite{XiChen2022FocalClickTP}}  & \scalebox{0.75}{COCO~\cite{TsungYiLin2014MicrosoftCC}+LVIS~\cite{AgrimGupta2019LVISAD}} & 31.7 & 0.05 (128) & 1.86  & 3.29  & 4.98 & 7.60 & 7.42 \\
   \multicolumn{2}{l|}{ FocalClick-SegFormerB0-S2~\cite{XiChen2022FocalClickTP}}  & \scalebox{0.75}{COCO~\cite{TsungYiLin2014MicrosoftCC}+LVIS~\cite{AgrimGupta2019LVISAD}}  & 14.8 & 0.10 (256) & 1.66 & 2.27 & 4.56 & 6.86 & 5.49 \\
   \multicolumn{2}{l|}{ FocalClick-SegFormerB3-S2~\cite{XiChen2022FocalClickTP}}  & \scalebox{0.75}{COCO~\cite{TsungYiLin2014MicrosoftCC}+LVIS~\cite{AgrimGupta2019LVISAD}}  & 46.1 & 0.30 (256) & 1.50 & \textbf{1.92} & 3.53 & 5.59 & \textbf{4.90} \\
   \multicolumn{2}{l|}{ PseudoClick-HRNet32~\cite{QinLiu2022PseudoClickII}}  & \scalebox{0.75}{COCO~\cite{TsungYiLin2014MicrosoftCC}+LVIS~\cite{AgrimGupta2019LVISAD}}  & -- & -- & 1.50 & 2.08 & -- & 5.54 & 5.11 \\
   \rowcolor{gray!10} 
   \multicolumn{2}{l|}{ InterFormer-Light }  & \scalebox{0.75}{COCO~\cite{TsungYiLin2014MicrosoftCC}+LVIS~\cite{AgrimGupta2019LVISAD}} & 8.9 & 0.23 (512) & 1.50 & 3.14 & 3.78 & 6.34 & 6.19 \\
   \rowcolor{gray!10} 
   \multicolumn{2}{l|}{ * InterFormer-Light }  & & 7.1 & 0.19 (512) & & & & & \\
   \rowcolor{gray!20} 
   \multicolumn{2}{l|}{ InterFormer-Tiny + Zoom-in}  & \scalebox{0.75}{COCO~\cite{TsungYiLin2014MicrosoftCC}+LVIS~\cite{AgrimGupta2019LVISAD}} & 15.9 & 0.42 (512) & 1.40 & 2.78 & 3.56 & 5.89 & 5.52 \\
   \rowcolor{gray!20} 
   \multicolumn{2}{l|}{ * InterFormer-Tiny + Zoom-in}  & & 9.1 & 0.24 (512) & & & & & \\
   \rowcolor{gray!30} 
   \multicolumn{2}{l|}{ InterFormer-Tiny}  & \scalebox{0.75}{COCO~\cite{TsungYiLin2014MicrosoftCC}+LVIS~\cite{AgrimGupta2019LVISAD}} & 18.9 & 0.50 (512) & \textbf{1.36} & 2.53 & \textbf{3.25} & \textbf{5.51} & 5.21 \\
   \rowcolor{gray!30} 
   \multicolumn{2}{l|}{ * InterFormer-Tiny}  & & 12.3 & 0.32 (512) & & & & & \\
   \hline
   \multicolumn{2}{l|}{SimpleClick-ViT-B~\cite{QinLiu2022SimpleClickII}}  & \scalebox{0.75}{COCO~\cite{TsungYiLin2014MicrosoftCC}+LVIS~\cite{AgrimGupta2019LVISAD}}  & 75.4 & 1.51 (448) & 1.48 & 1.97 & 3.43 & 5.62 & 5.06 \\
   \multicolumn{2}{l|}{SimpleClick-ViT-L~\cite{QinLiu2022SimpleClickII}}  & \scalebox{0.75}{COCO~\cite{TsungYiLin2014MicrosoftCC}+LVIS~\cite{AgrimGupta2019LVISAD}}  & 165.7 & 3.33 (448) & 1.40 & 1.89 & 2.95 & 4.89 & 4.81 \\
   \multicolumn{2}{l|}{SimpleClick-ViT-H~\cite{QinLiu2022SimpleClickII}}  & \scalebox{0.75}{COCO~\cite{TsungYiLin2014MicrosoftCC}+LVIS~\cite{AgrimGupta2019LVISAD}}  & 386.8 & 7.76 (448) & 1.50 & 1.75 & 2.85 & 4.70 & 4.78 \\
   \bottomrule[1pt]
   \end{tabular}
   }
   \end{center}
   \vspace{-0.5em}
   \caption{Evaluation results of InterFormer on GrabCut, Berkeley, SBD, and DAVIS datasets. InferFormer's SPC and PIE are measured by averaging inference time across 20 clicks, accounting for image preprocessing. $*$ indicates that SPC and PIE are evaluated during interaction without considering image preprocessing. These results demonstrate the potential of InterFormer to improve segmentation performance and efficiency in interactive segmentation.}
   \label{tab:mainresults}
   \vspace{-1em}
   \end{table*}

\textbf{Model series.} 
We propose two configurations of the InterFormer model shown in Table~\ref{tab:configs}, which can be deployed on CPU-only devices. Both configurations use a light architecture, with 32, 64, 128, and 256 channels in the four stages of I-MSA. Additionally, the UperNet model has 64 channels. To further reduce computational complexity, we apply a zoom-in strategy, but only to the InterFormer-Tiny configuration due to its larger architecture. Specifically, we apply the zoom-in strategy to the deeper blocks at each stage, starting from the $2$nd, $2$nd, $3$rd and $2$nd blocks respectively in the four stages. The shallower blocks of InterFormer's I-MSA process the global features without using the zoom-in strategy.

\textbf{Training strategy.} 
Our models are trained using a combination of the COCO~\cite{TsungYiLin2014MicrosoftCC} and LVIS~\cite{AgrimGupta2019LVISAD}, following the previous works~\cite{XiChen2022FocalClickTP,QinLiu2022PseudoClickII}. We utilize widely adopted data augmentation techniques, including random resizing with a scale between 0.5 and 2.0, random flipping, random cropping, and color jittering. Each augmented image is padded to a final size of $512\times512$ pixels. As our ViT backbones are pre-trained on $224\times224$ pixel images by MAE~\cite{KaimingHe2021MaskedAA}, we resize the pre-trained absolute positional embeddings to match the size of our images. During training, our models are allowed to perform up to 20 pre-interactions, simulating clicks as model inputs. We use a probability decay ratio of $\gamma = 0.6$ to sample the number of simulations. Our models are trained using a batch size of 16 for 40k iterations in ablation studies, and 320k iterations for main experiments. We use the AdamW optimizer with a learning rate of $1\times10^{-4}$ and a polynomial decay strategy, setting the layer decay rate to $0.65$ for ViT-Base and $0.75$ for ViT-Large. We do not use any zoom-in strategies during training.

\textbf{Evaluation strategy.} 
Each evaluation image is padded to a multiple of the maximum stride of InterFormer's internal features (\ie 32). The positional embeddings of ViTs are dynamically resized based on the input image size to match with the padded images. Following the previous works~\cite{KonstantinSofiiuk2021RevivingIT, XiChen2022FocalClickTP}, we employ InterFormer to iteratively perform interactive segmentation with the maximum click number set to $20$. Each click is placed in the center of the erroneous region. Additionally, we utilize the zoom-in strategy to evaluate the performance of InterFormer-Tiny with faster model inference.

\textbf{Evaluation metrics.}
We report NoC IoU (Number of Clicks) of each method, which indicates the average number of clicks required to achieve the desired IoU. In addition, we
focus on evaluating the speed of InterFormer on CPU-only devices to provide insights into the actual performance of interactive segmentation models.

\begin{figure*}[t]
   \begin{center}
      \includegraphics[width=1.0\linewidth]{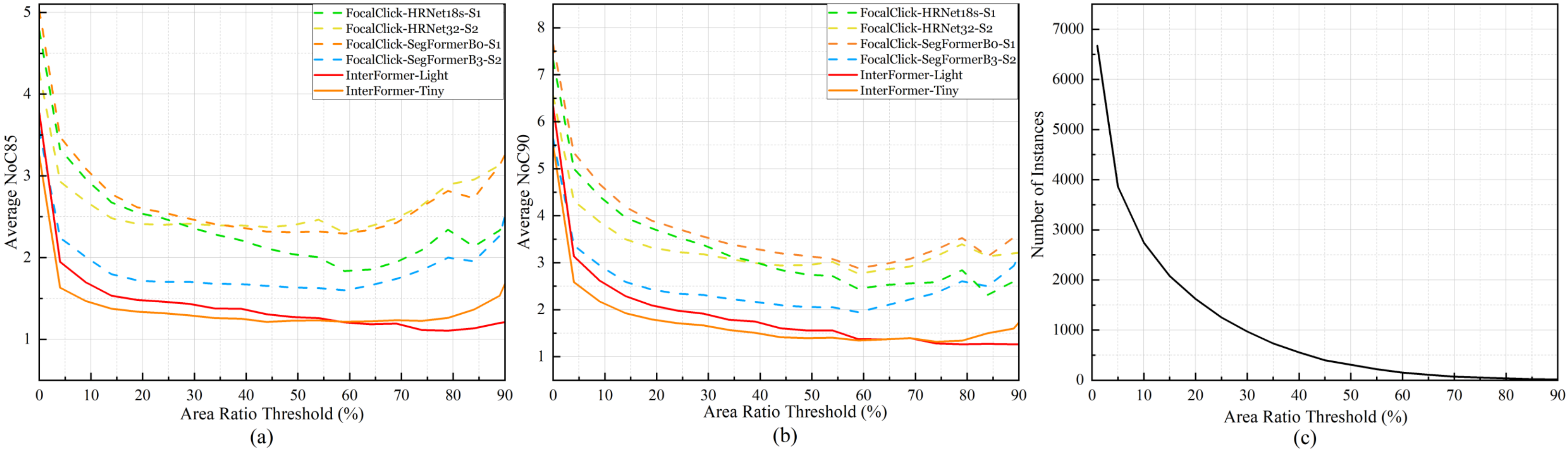}
   \end{center}
      \vspace{-1em}
      \caption{Comparison of InterFormer and FocalClick on objects of varying sizes, measured by their average NoC85 (a) and NoC90 (b) for objects with area ratios larger than different thresholds. The larger thresholds (\eg $> 80\%$) filter out more instances (c), resulting in the noisy estimation of the NoC metrics. Such results indicate that InterFormer exhibits superior performance over FocalClick, particularly for larger objects, demonstrating its greater robustness and versatility.}
   \label{fig:varyingsizesobjects}
\end{figure*}

To evaluate InterFormer's speed, we adopt the widely used metric, Seconds Per Click (SPC). However, each interactive segmentation model has its own resizing strategy, resulting in inconsistent SPC evaluations. For instance, FocalClick~\cite{XiChen2022FocalClickTP} downsamples images into $256\times256$ size to obtain coarse segmentation results, which were then refined using additional modules. To address the inconsistent evaluation, we propose a novel metric called Pixel-wise Inference Efficiency (PIE), which measures a model's efficiency in pixel-wise segmentation across potentially large or small objects. We evaluate PIE by calculating the pixel-wise average of SPC.

\begin{figure}[ht]
   \begin{center}
      \includegraphics[width=0.8\linewidth]{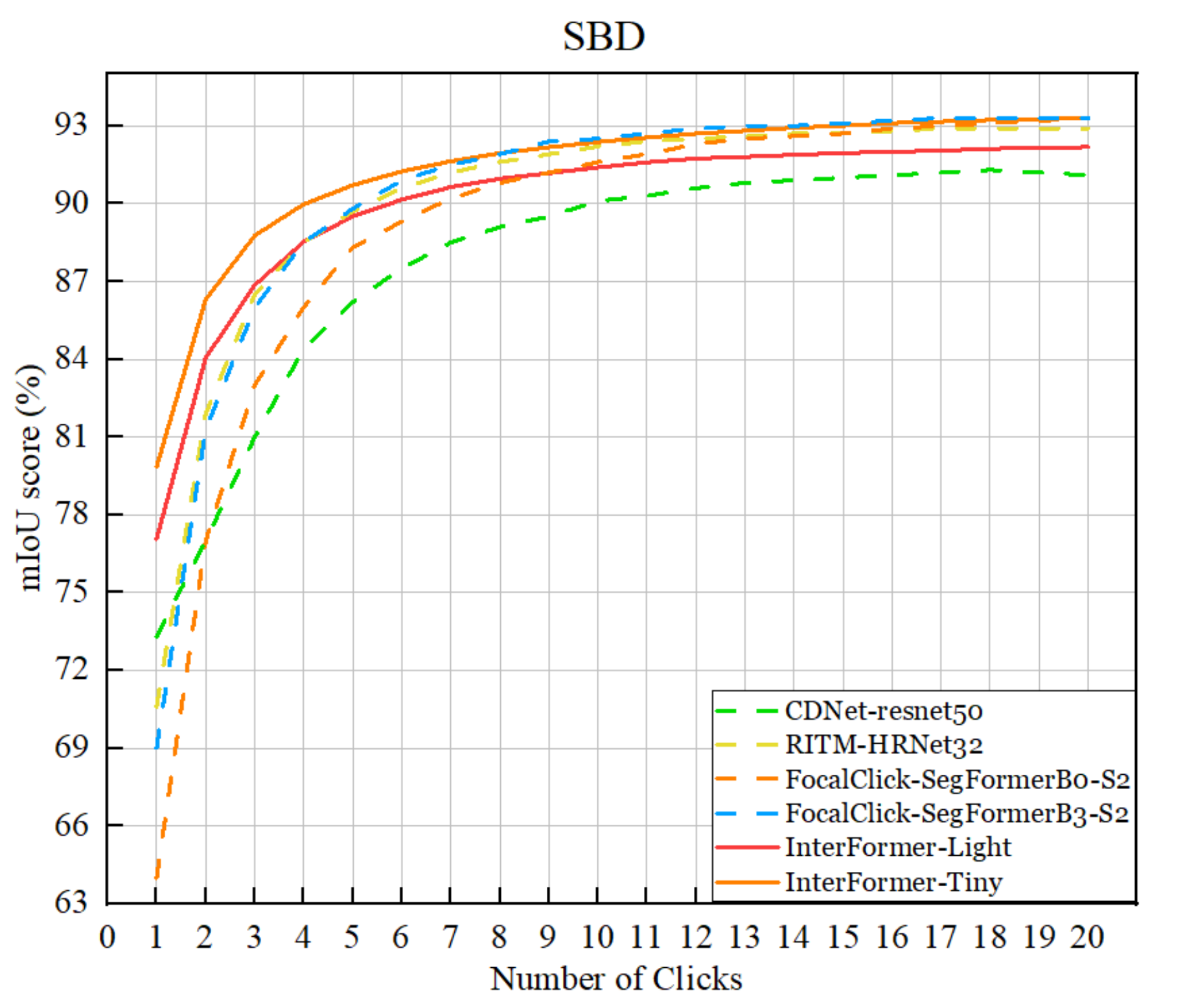}
   \end{center}
   \vspace{-1em}
      \caption{Convergence analysis of interactive segmentation models. It is worth noting that our InterFormer manifests exceptional convergence, outpacing other recently proposed methods of similar computational demands, by achieving $90\%$ IoU within just four clicks.}
\label{fig:convergenceanalysis}
\end{figure}

\begin{table}[ht]
   \centering
   \begin{tabular}{ccc}
      \toprule
      \textbf{Model} & \textbf{Training} & \textbf{NoC85/NoC90} \\
      \midrule
      ViT-B-FPN & RITM  & 7.53/10.70 \\
      ViT-B-I-MSA-Light & RITM  & 5.38/8.27 \\
      ViT-B-I-MSA-Light & Simple & 5.35/8.34 \\
      VIT-L-I-MSA-Light & Simple & 4.80/7.71 \\
      \bottomrule
      \end{tabular}%
   \vspace{1em}
     \caption{Ablation study on InterFormer. We find that larger backbones, such as ViT-Large, significantly improve model performance compared to ViT-Base, and the interactive module plays a more crucial role in enhancing the performance of the interactive segmentation model. Instead, the choice of training strategy has a slight impact on the performance.}
   \label{tab:ablationstudy}%
 \end{table}%

\subsection{Main Result}\label{sec:mainresult}

\textbf{Performance on existing benchmarks.}
We present the results of InterFormer, as shown in Table~\ref{sec:mainresult}. To facilitate comparison, we categorize the previous methods into different blocks based on their training data or PIE, as larger training datasets and computation trivially leads to improved performance. Among the methods that allow real-time interaction on CPU-only devices, our proposed InterFormer-Tiny achieves state-of-the-art performance on the largest validation dataset, SBD~\cite{BharathHariharan2011SemanticCF}, consisting of 6671 images, for both NoC85 and NoC90. Additionally, InterFormer exhibits competitive performance on other datasets. However, it is noteworthy that FocalClick~\cite{XiChen2022FocalClickTP} outperforms InterFormer on the DAVIS dataset due to its significantly larger computations (measured by PIE). In addition, InterFormer showcases superior efficiency with generally the lowest PIE, indicating its high efficiency in interactive segmentation on CPU-only devices, irrespective of the varying image and object sizes.

\textbf{Convergence Analysis.}
In Figure~\ref{fig:convergenceanalysis}, we present a comprehensive analysis of the convergence of various recently proposed methods. Notably, our InterFormer outperforms other state-of-the-art methods with similar computational complexity, reaching $90\%$ IoU within merely four initial clicks. This efficiency is largely owed to the potent generalization features of the MAE-pretrained ViTs used by InterFormer.

\textbf{Evaluation on differently sized objects.}
To showcase the consistent performance of InterFormer regardless of the sizes of objects of interest, we conduct experiments on objects of varying sizes. In Figure~\ref{fig:varyingsizesobjects}, we present the average NoC85 and NoC90 of each model for objects with area ratios larger than the different thresholds (x-axis). We compare InterFormer with FocalClick that has a low SPC (as reported in Table~\ref{tab:mainresults}) and a resizing strategy that may degenerate in the case of large objects. The results demonstrate that InterFormer outperforms FocalClick in terms of NoC85 and NoC90 over large objects, as depicted in Figure~\ref{fig:varyingsizesobjects}. Moreover, InterFormer maintains its consistency in performance across the different size thresholds, further validating its robustness and versatility.

Besides, our findings highlight the importance of using PIE as a metric for evaluating interactive segmentation models in practical situations. There is no universal resizing strategy that can work well for objects of all sizes. Therefore, using a metric that considers the performance of a model across different sizes is crucial in evaluating its effectiveness. InterFormer's ability to maintain its performance across different object sizes underscores its utility in real-world applications.

\subsection{Ablation Study}\label{sec:ablationstudy}

We conducted the ablation studies to evaluate the impact of various model components on the performance of InterFormer. Specifically, we replaced the proposed I-MSA module with a vanilla FPN-like module where click embeddings are added directly to the features extracted by the large encoder. We also evaluated the performance of models trained on the RITM training strategy~\cite{KonstantinSofiiuk2021RevivingIT} and our simpler training strategy, as well as the impact of using a larger backbone such as ViT-Large. To expedite the analysis process, we trained each model for a shorter duration of $40k$ iterations instead of the standard $320k$ iterations. 

Table~\ref{tab:ablationstudy} presents the results of our comprehensive ablation study. Our findings reveal that the choice of training strategy has a minimal impact on the performance of InterFormer, whereas using a larger backbone like ViT-Large significantly improves the model's performance. Notably, the study highlights the critical role of the interactive module in improving the performance of the interactive segmentation model in our pipeline. Specifically, our findings show that using a vanilla FPN-like interactive module leads to a deterioration in performance. However, our proposed I-MSA module significantly outperforms the vanilla module, indicating its effectiveness in improving the performance of the interactive segmentation model.

\subsection{Qualitative Result}\label{sec:qualitativeresult}
\begin{figure}[t]
   \begin{center}
      \includegraphics[width=1.0\linewidth]{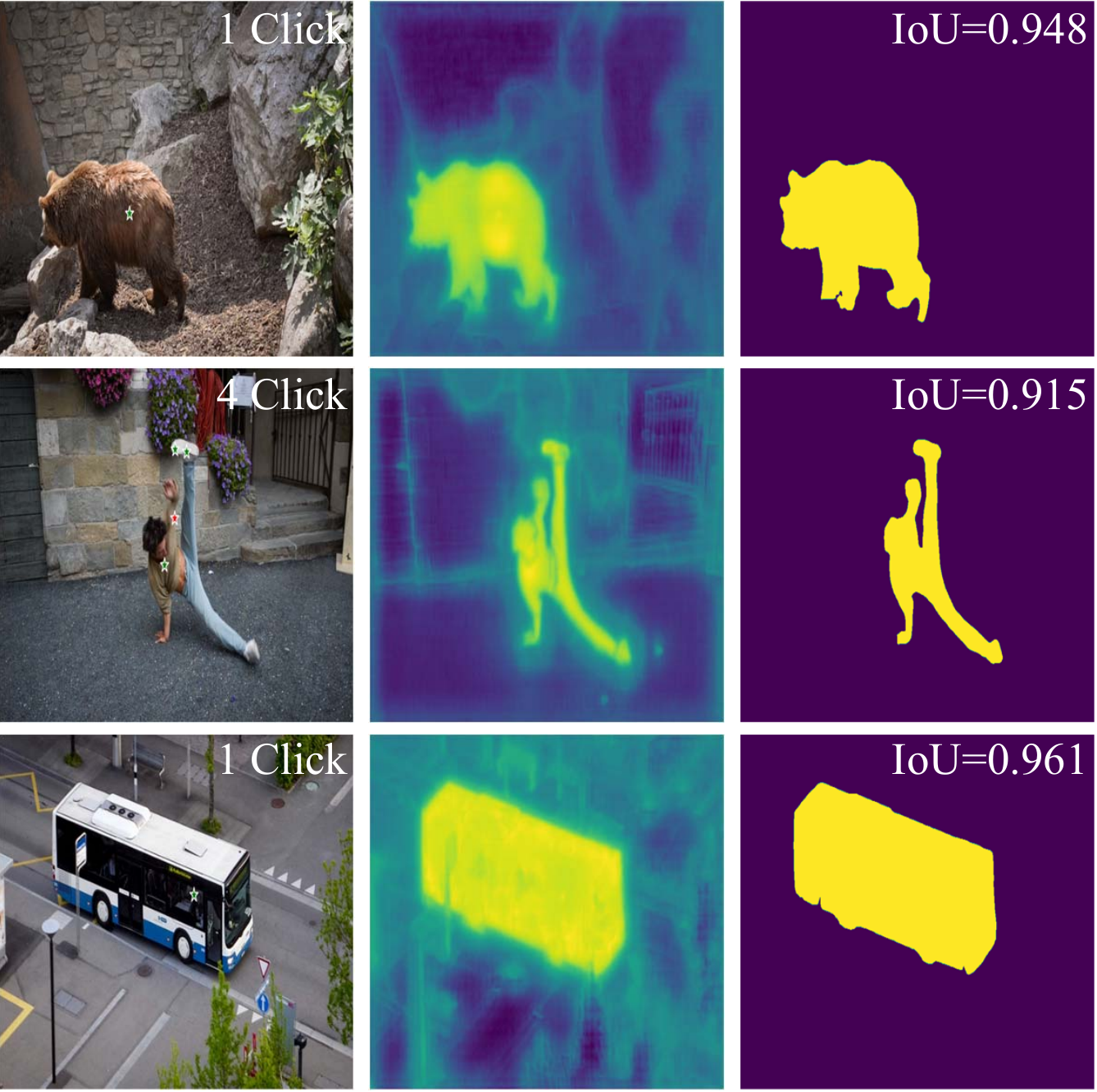}
   \end{center}
   \vspace{-1em}
      \caption{Qualitative results on DAVIS. The first column displays the original image with clicks, denoted by pentagrams (green for positive clicks and red for negative). The second column presents the model's predicted logits illustrated as a heatmap. The third column showcases the predicted segmentation results, with the IoU calculated against the ground truth indicated within.}
   \label{fig:qualitative}
   \end{figure}

We performed a qualitative assessment of InterFormer-Tiny trained for $320k$ iterations. The outcomes of this evaluation are depicted in Figure~\ref{fig:qualitative}, where InterFormer can generate a IoU score exceeding 0.9 for two out of the three cases with only one click.

\section{Conclusion}\label{sec:conclusion}
This paper focuses on improving interactive efficiency and speed of interactive image segmentation. Previous methods often compute image features repeatedly at each interaction step. In this paper, we propose InterFormer, a new pipeline that separates image processing from interactive segmentation. InterFormer uses a large vision transformer (ViT) on high-performance devices to preprocess images in parallel and then employs a lightweight interactive multi-head self-attention (I-MSA) module on low-power devices for real-time segmentation. Extensive experiments demonstrate that InterFormer achieves high-quality interactive segmentation on CPU-only devices while outperforming previous models in terms of efficiency and quality.

\paragraph{Acknowledgements.}
This work was supported by National Key R\&D Program of China (No.2022ZD0118202), the National Science Fund for Distinguished Young Scholars (No.62025603), the National Natural Science Foundation of China (No. U21B2037, No. U22B2051, No. 62176222, No. 62176223, No. 62176226, No. 62072386, No. 62072387, No. 62072389, No. 62002305 and No. 62272401), and the Natural Science Foundation of Fujian Province of China (No.2021J01002,  No.2022J06001).

{\small
\bibliographystyle{ieee_fullname}
\bibliography{camera-ready-interformer}

\begin{thebibliography}{10}\itemsep=-1pt

\bibitem{DavidAcuna2018EfficientIA}
David Acuna, Huan Ling, Amlan Kar, and Sanja Fidler.
\newblock Efficient interactive annotation of segmentation datasets with
  polygon-rnn++.
\newblock In {\em 2018 {IEEE} Conference on Computer Vision and Pattern
  Recognition, {CVPR} 2018, Salt Lake City, UT, USA, June 18-22, 2018}, pages
  859--868. {IEEE} Computer Society, 2018.

\bibitem{EirikurAgustsson2018InteractiveFI}
Eirikur Agustsson, Jasper R.~R. Uijlings, and Vittorio Ferrari.
\newblock Interactive full image segmentation by considering all regions
  jointly.
\newblock In {\em {IEEE} Conference on Computer Vision and Pattern Recognition,
  {CVPR} 2019, Long Beach, CA, USA, June 16-20, 2019}, pages 11622--11631.
  Computer Vision Foundation / {IEEE}, 2019.

\bibitem{SaberMirzaeeBafti2021ACS}
Saber~Mirzaee Bafti, Chee~Siang Ang, Md.~Moinul Hossain, Gianluca Marcelli,
  Marc Alemany-Fornes, and Anastasios~D. Tsaousis.
\newblock A crowdsourcing semi-automatic image segmentation platform for cell
  biology.
\newblock 2021.

\bibitem{JunjieBai2014ErrorTolerantSB}
Junjie Bai and Xiaodong Wu.
\newblock Error-tolerant scribbles based interactive image segmentation.
\newblock In {\em 2014 {IEEE} Conference on Computer Vision and Pattern
  Recognition, {CVPR} 2014, Columbus, OH, USA, June 23-28, 2014}, pages
  392--399. {IEEE} Computer Society, 2014.

\bibitem{RodrigoBenenson2019LargescaleIO}
Rodrigo Benenson, Stefan Popov, and Vittorio Ferrari.
\newblock Large-scale interactive object segmentation with human annotators.
\newblock In {\em {IEEE} Conference on Computer Vision and Pattern Recognition,
  {CVPR} 2019, Long Beach, CA, USA, June 16-20, 2019}, pages 11700--11709.
  Computer Vision Foundation / {IEEE}, 2019.

\bibitem{GedasBertasius2019ClassifyingSA}
Gedas Bertasius and Lorenzo Torresani.
\newblock Classifying, segmenting, and tracking object instances in video with
  mask propagation.
\newblock In {\em 2020 {IEEE/CVF} Conference on Computer Vision and Pattern
  Recognition, {CVPR} 2020, Seattle, WA, USA, June 13-19, 2020}, pages
  9736--9745. {IEEE}, 2020.

\bibitem{YuriBoykov2001InteractiveGC}
Yuri~Y Boykov and M-P Jolly.
\newblock Interactive graph cuts for optimal boundary \& region segmentation of
  objects in nd images.
\newblock 1:105--112, 2001.

\bibitem{HolgerCaesar2019nuScenesAM}
Holger Caesar, Varun Bankiti, Alex~H Lang, Sourabh Vora, Venice~Erin Liong,
  Qiang Xu, Anush Krishnan, Yu Pan, Giancarlo Baldan, and Oscar Beijbom.
\newblock nuscenes: A multimodal dataset for autonomous driving.
\newblock In {\em Proceedings of the IEEE/CVF conference on computer vision and
  pattern recognition}, pages 11621--11631, 2020.

\bibitem{XiChen2021ConditionalDF}
Xi Chen, Zhiyan Zhao, Feiwu Yu, Yilei Zhang, and Manni Duan.
\newblock Conditional diffusion for interactive segmentation.
\newblock pages 7345--7354, 2021.

\bibitem{XiChen2022FocalClickTP}
Xi Chen, Zhiyan Zhao, Yilei Zhang, Manni Duan, Donglian Qi, and Hengshuang
  Zhao.
\newblock Focalclick: towards practical interactive image segmentation.
\newblock pages 1300--1309, 2022.

\bibitem{DengxinDai2014TheSO}
Dengxin Dai, Hayko Riemenschneider, and Luc~Van Gool.
\newblock The synthesizability of texture examples.
\newblock In {\em 2014 {IEEE} Conference on Computer Vision and Pattern
  Recognition, {CVPR} 2014, Columbus, OH, USA, June 23-28, 2014}, pages
  3027--3034. {IEEE} Computer Society, 2014.

\bibitem{AlexeyDosovitskiy2020AnII}
Alexey Dosovitskiy, Lucas Beyer, Alexander Kolesnikov, Dirk Weissenborn,
  Xiaohua Zhai, Thomas Unterthiner, Mostafa Dehghani, Matthias Minderer, Georg
  Heigold, Sylvain Gelly, Jakob Uszkoreit, and Neil Houlsby.
\newblock An image is worth 16x16 words: Transformers for image recognition at
  scale.
\newblock In {\em 9th International Conference on Learning Representations,
  {ICLR} 2021, Virtual Event, Austria, May 3-7, 2021}. OpenReview.net, 2021.

\bibitem{MarkEveringham2010ThePV}
Mark Everingham, Luc~Van Gool, Christopher Williams, John Winn, and Andrew
  Zisserman.
\newblock The pascal visual object classes (voc) challenge.
\newblock {\em International Journal of Computer Vision}, 2010.

\bibitem{HaoqiFan2021MultiscaleVT}
Haoqi Fan, Bo Xiong, Karttikeya Mangalam, Yanghao Li, Zhicheng Yan, Jitendra
  Malik, and Christoph Feichtenhofer.
\newblock Multiscale vision transformers.
\newblock pages 6824--6835, 2021.

\bibitem{MarcoForte2020GettingT9}
Marco Forte, Brian Price, Scott Cohen, Ning Xu, and François Piti{\'e}.
\newblock Getting to 99\% accuracy in interactive segmentation.
\newblock {\em arXiv: Computer Vision and Pattern Recognition}, 2020.

\bibitem{LeoGrady2006RandomWF}
Leo Grady.
\newblock Random walks for image segmentation.
\newblock {\em IEEE Transactions on Pattern Analysis and Machine Intelligence},
  2006.

\bibitem{VarunGulshan2010GeodesicSC}
Varun Gulshan, Carsten Rother, Antonio Criminisi, Andrew Blake, and Andrew
  Zisserman.
\newblock Geodesic star convexity for interactive image segmentation.
\newblock In {\em The Twenty-Third {IEEE} Conference on Computer Vision and
  Pattern Recognition, {CVPR} 2010, San Francisco, CA, USA, 13-18 June 2010},
  pages 3129--3136. {IEEE} Computer Society, 2010.

\bibitem{AgrimGupta2019LVISAD}
Agrim Gupta, Piotr Doll{\'{a}}r, and Ross~B. Girshick.
\newblock {LVIS:} {A} dataset for large vocabulary instance segmentation.
\newblock In {\em {IEEE} Conference on Computer Vision and Pattern Recognition,
  {CVPR} 2019, Long Beach, CA, USA, June 16-20, 2019}, pages 5356--5364.
  Computer Vision Foundation / {IEEE}, 2019.

\bibitem{YuyingHao2021EdgeFlowAP}
Yuying Hao, Yi Liu, Zewu Wu, Lin Han, Yizhou Chen, Guowei Chen, Lutao Chu,
  Shiyu Tang, Zhiliang Yu, Zeyu Chen, et~al.
\newblock Edgeflow: Achieving practical interactive segmentation with
  edge-guided flow.
\newblock pages 1551--1560, 2021.

\bibitem{BharathHariharan2011SemanticCF}
Bharath Hariharan, Pablo Arbelaez, Lubomir~D. Bourdev, Subhransu Maji, and
  Jitendra Malik.
\newblock Semantic contours from inverse detectors.
\newblock In Dimitris~N. Metaxas, Long Quan, Alberto Sanfeliu, and Luc~Van
  Gool, editors, {\em {IEEE} International Conference on Computer Vision,
  {ICCV} 2011, Barcelona, Spain, November 6-13, 2011}, pages 991--998. {IEEE}
  Computer Society, 2011.

\bibitem{KaimingHe2021MaskedAA}
Kaiming He, Xinlei Chen, Saining Xie, Yanghao Li, Piotr Doll{\'a}r, and Ross
  Girshick.
\newblock Masked autoencoders are scalable vision learners.
\newblock {\em arXiv: Computer Vision and Pattern Recognition}, 2021.

\bibitem{KaimingHe2017MaskR}
Kaiming He, Georgia Gkioxari, Piotr Doll{\'{a}}r, and Ross~B. Girshick.
\newblock Mask {R-CNN}.
\newblock In {\em {IEEE} International Conference on Computer Vision, {ICCV}
  2017, Venice, Italy, October 22-29, 2017}, pages 2980--2988. {IEEE} Computer
  Society, 2017.

\bibitem{hu2021istr}
Jie Hu, Liujuan Cao, Yao Lu, ShengChuan Zhang, Yan Wang, Ke Li, Feiyue Huang,
  Ling Shao, and Rongrong Ji.
\newblock Istr: End-to-end instance segmentation with transformers.
\newblock {\em arXiv preprint arXiv:2105.00637}, 2021.

\bibitem{hu2023you}
Jie Hu, Linyan Huang, Tianhe Ren, Shengchuan Zhang, Rongrong Ji, and Liujuan
  Cao.
\newblock You only segment once: Towards real-time panoptic segmentation.
\newblock In {\em Proceedings of the IEEE/CVF Conference on Computer Vision and
  Pattern Recognition}, pages 17819--17829, 2023.

\bibitem{JangWonDong2019InteractiveIS}
Won{-}Dong Jang and Chang{-}Su Kim.
\newblock Interactive image segmentation via backpropagating refinement scheme.
\newblock In {\em {IEEE} Conference on Computer Vision and Pattern Recognition,
  {CVPR} 2019, Long Beach, CA, USA, June 16-20, 2019}, pages 5297--5306.
  Computer Vision Foundation / {IEEE}, 2019.

\bibitem{KevinMcGuinness2010ACE}
Noel E.~O'Connor Kevin~McGuinness.
\newblock A comparative evaluation of interactive segmentation algorithms.
\newblock {\em Pattern Recognition}, 2010.

\bibitem{KonstantinSofiiuk2021RevivingIT}
Anton~Konushin Konstantin~Sofiiuk, Ilia A.~Petrov.
\newblock Reviving iterative training with mask guidance for interactive
  segmentation.
\newblock {\em arXiv: Computer Vision and Pattern Recognition}, 2021.

\bibitem{YanghaoLiExploringPV}
Yanghao Li, Hanzi Mao, Ross Girshick, and Kaiming He.
\newblock Exploring plain vision transformer backbones for object detection.

\bibitem{ZhuwenLi2018InteractiveIS}
Zhuwen Li, Qifeng Chen, and Vladlen Koltun.
\newblock Interactive image segmentation with latent diversity.
\newblock In {\em 2018 {IEEE} Conference on Computer Vision and Pattern
  Recognition, {CVPR} 2018, Salt Lake City, UT, USA, June 18-22, 2018}, pages
  577--585. {IEEE} Computer Society, 2018.

\bibitem{JunHaoLiew2021DeepIT}
Jun~Hao Liew, Scott Cohen, Brian Price, Long Mai, and Jiashi Feng.
\newblock Deep interactive thin object selection.
\newblock pages 305--314, 2021.

\bibitem{TsungYiLin2016FeaturePN}
Tsung{-}Yi Lin, Piotr Doll{\'{a}}r, Ross~B. Girshick, Kaiming He, Bharath
  Hariharan, and Serge~J. Belongie.
\newblock Feature pyramid networks for object detection.
\newblock In {\em 2017 {IEEE} Conference on Computer Vision and Pattern
  Recognition, {CVPR} 2017, Honolulu, HI, USA, July 21-26, 2017}, pages
  936--944. {IEEE} Computer Society, 2017.

\bibitem{TsungYiLin2014MicrosoftCC}
Tsung-Yi Lin, Michael Maire, Serge Belongie, James Hays, Pietro Perona, Deva
  Ramanan, Piotr Doll{\'a}r, and C.~Lawrence Zitnick.
\newblock Microsoft coco: Common objects in context.
\newblock {\em Lecture Notes in Computer Science}, 2014.

\bibitem{ZhengLinFocusCutDI}
Zheng Lin, Zheng-Peng Duan, Zhao Zhang, Chun-Le Guo, and Ming-Ming Cheng.
\newblock Focuscut: Diving into a focus view in interactive segmentation.
\newblock pages 2637--2646, 2022.

\bibitem{ZhengLin2020InteractiveIS}
Zheng Lin, Zhao Zhang, Lin{-}Zhuo Chen, Ming{-}Ming Cheng, and Shao{-}Ping Lu.
\newblock Interactive image segmentation with first click attention.
\newblock In {\em 2020 {IEEE/CVF} Conference on Computer Vision and Pattern
  Recognition, {CVPR} 2020, Seattle, WA, USA, June 13-19, 2020}, pages
  13336--13345. {IEEE}, 2020.

\bibitem{HuanLing2019FastIO}
Huan Ling, Jun Gao, Amlan Kar, Wenzheng Chen, and Sanja Fidler.
\newblock Fast interactive object annotation with curve-gcn.
\newblock In {\em {IEEE} Conference on Computer Vision and Pattern Recognition,
  {CVPR} 2019, Long Beach, CA, USA, June 16-20, 2019}, pages 5257--5266.
  Computer Vision Foundation / {IEEE}, 2019.

\bibitem{GeertLitjens2017ASO}
Geert Litjens, Thijs Kooi, Babak~Ehteshami Bejnordi, Arnaud Arindra~Adiyoso
  Setio, Francesco Ciompi, Mohsen Ghafoorian, Jeroen~Awm Van Der~Laak, Bram
  Van~Ginneken, and Clara~I S{\'a}nchez.
\newblock A survey on deep learning in medical image analysis.
\newblock {\em Medical image analysis}, 42:60--88, 2017.

\bibitem{QinLiu2022SimpleClickII}
Qin Liu, Zhenlin Xu, Gedas Bertasius, and Marc Niethammer.
\newblock Simpleclick: Interactive image segmentation with simple vision
  transformers.
\newblock {\em arXiv preprint arXiv:2210.11006}, 2022.

\bibitem{QinLiu2022PseudoClickII}
Qin Liu, Meng Zheng, Benjamin Planche, Srikrishna Karanam, Terrence Chen, Marc
  Niethammer, and Ziyan Wu.
\newblock Pseudoclick: Interactive image segmentation with click imitation.
\newblock pages 728--745, 2022.

\bibitem{liu2021swin}
Ze Liu, Yutong Lin, Yue Cao, Han Hu, Yixuan Wei, Zheng Zhang, Stephen Lin, and
  Baining Guo.
\newblock Swin transformer: Hierarchical vision transformer using shifted
  windows.
\newblock In {\em Proceedings of the IEEE/CVF international conference on
  computer vision}, pages 10012--10022, 2021.

\bibitem{MajumderSoumajit2019ContentAwareMG}
Soumajit Majumder and Angela Yao.
\newblock Content-aware multi-level guidance for interactive instance
  segmentation.
\newblock In {\em Proceedings of the IEEE/CVF Conference on Computer Vision and
  Pattern Recognition}, pages 11602--11611, 2019.

\bibitem{KevisKokitsiManinis2017DeepEC}
Kevis{-}Kokitsi Maninis, Sergi Caelles, Jordi Pont{-}Tuset, and Luc~Van Gool.
\newblock Deep extreme cut: From extreme points to object segmentation.
\newblock In {\em 2018 {IEEE} Conference on Computer Vision and Pattern
  Recognition, {CVPR} 2018, Salt Lake City, UT, USA, June 18-22, 2018}, pages
  616--625. {IEEE} Computer Society, 2018.

\bibitem{FedericoPerazzi2016ABD}
Federico Perazzi, Jordi Pont{-}Tuset, Brian McWilliams, Luc~Van Gool, Markus~H.
  Gross, and Alexander Sorkine{-}Hornung.
\newblock A benchmark dataset and evaluation methodology for video object
  segmentation.
\newblock In {\em 2016 {IEEE} Conference on Computer Vision and Pattern
  Recognition, {CVPR} 2016, Las Vegas, NV, USA, June 27-30, 2016}, pages
  724--732. {IEEE} Computer Society, 2016.

\bibitem{CarstenRother2004GrabCutIF}
Carsten Rother, Vladimir Kolmogorov, and Andrew Blake.
\newblock "grabcut" interactive foreground extraction using iterated graph
  cuts.
\newblock {\em ACM transactions on graphics (TOG)}, 23(3):309--314, 2004.

\bibitem{KonstantinSofiiuk2020fBRSRB}
Konstantin Sofiiuk, Ilia~A. Petrov, Olga Barinova, and Anton Konushin.
\newblock {F-BRS:} rethinking backpropagating refinement for interactive
  segmentation.
\newblock In {\em 2020 {IEEE/CVF} Conference on Computer Vision and Pattern
  Recognition, {CVPR} 2020, Seattle, WA, USA, June 13-19, 2020}, pages
  8620--8629. {IEEE}, 2020.

\bibitem{RobinStrudel2021SegmenterTF}
Robin Strudel, Ricardo Garcia, Ivan Laptev, and Cordelia Schmid.
\newblock Segmenter: Transformer for semantic segmentation.
\newblock pages 7262--7272, 2021.

\bibitem{ChenSun2017RevisitingUE}
Chen Sun, Abhinav Shrivastava, Saurabh Singh, and Abhinav Gupta.
\newblock Revisiting unreasonable effectiveness of data in deep learning era.
\newblock In {\em {IEEE} International Conference on Computer Vision, {ICCV}
  2017, Venice, Italy, October 22-29, 2017}, pages 843--852. {IEEE} Computer
  Society, 2017.

\bibitem{HugoTouvron2020TrainingDI}
Hugo Touvron, Matthieu Cord, Matthijs Douze, Francisco Massa, Alexandre
  Sablayrolles, and Herv{\'{e}} J{\'{e}}gou.
\newblock Training data-efficient image transformers {\&} distillation through
  attention.
\newblock In Marina Meila and Tong Zhang, editors, {\em Proceedings of the 38th
  International Conference on Machine Learning, {ICML} 2021, 18-24 July 2021,
  Virtual Event}, volume 139 of {\em Proceedings of Machine Learning Research},
  pages 10347--10357. {PMLR}, 2021.

\bibitem{AshishVaswani2017AttentionIA}
Ashish Vaswani, Noam Shazeer, Niki Parmar, Jakob Uszkoreit, Llion Jones,
  Aidan~N. Gomez, Lukasz Kaiser, and Illia Polosukhin.
\newblock Attention is all you need.
\newblock In Isabelle Guyon, Ulrike von Luxburg, Samy Bengio, Hanna~M. Wallach,
  Rob Fergus, S.~V.~N. Vishwanathan, and Roman Garnett, editors, {\em Advances
  in Neural Information Processing Systems 30: Annual Conference on Neural
  Information Processing Systems 2017, December 4-9, 2017, Long Beach, CA,
  {USA}}, pages 5998--6008, 2017.

\bibitem{WenhaiWang2021PyramidVT}
Wenhai Wang, Enze Xie, Xiang Li, Deng-Ping Fan, Kaitao Song, Ding Liang, Tong
  Lu, Ping Luo, and Ling Shao.
\newblock Pyramid vision transformer: A versatile backbone for dense prediction
  without convolutions.
\newblock pages 568--578, 2021.

\bibitem{JiajunWu2014MILCutAS}
Jiajun Wu, Yibiao Zhao, Jun{-}Yan Zhu, Siwei Luo, and Zhuowen Tu.
\newblock Milcut: {A} sweeping line multiple instance learning paradigm for
  interactive image segmentation.
\newblock In {\em 2014 {IEEE} Conference on Computer Vision and Pattern
  Recognition, {CVPR} 2014, Columbus, OH, USA, June 23-28, 2014}, pages
  256--263. {IEEE} Computer Society, 2014.

\bibitem{YuHuanWu2021P2TPP}
Yu-Huan Wu, Yun Liu, Xin Zhan, and Ming-Ming Cheng.
\newblock P2t: Pyramid pooling transformer for scene understanding.
\newblock {\em IEEE Transactions on Pattern Analysis and Machine Intelligence},
  2022.

\bibitem{TeteXiao2018UnifiedPP}
Tete Xiao, Yingcheng Liu, Bolei Zhou, Yuning Jiang, and Jian Sun.
\newblock Unified perceptual parsing for scene understanding.
\newblock pages 418--434, 2018.

\bibitem{EnzeXie2021SegFormerSA}
Enze Xie, Wenhai Wang, Zhiding Yu, Anima Anandkumar, Jose~M Alvarez, and Ping
  Luo.
\newblock Segformer: Simple and efficient design for semantic segmentation with
  transformers.
\newblock {\em Advances in Neural Information Processing Systems},
  34:12077--12090, 2021.

\bibitem{NingXu2017DeepIM}
Ning Xu, Brian~L. Price, Scott Cohen, and Thomas~S. Huang.
\newblock Deep image matting.
\newblock In {\em 2017 {IEEE} Conference on Computer Vision and Pattern
  Recognition, {CVPR} 2017, Honolulu, HI, USA, July 21-26, 2017}, pages
  311--320. {IEEE} Computer Society, 2017.

\bibitem{NingXu2016DeepIO}
Ning Xu, Brian~L. Price, Scott Cohen, Jimei Yang, and Thomas~S. Huang.
\newblock Deep interactive object selection.
\newblock In {\em 2016 {IEEE} Conference on Computer Vision and Pattern
  Recognition, {CVPR} 2016, Las Vegas, NV, USA, June 27-30, 2016}, pages
  373--381. {IEEE} Computer Society, 2016.

\bibitem{LiYuan2021TokenstoTokenVT}
Li Yuan, Yunpeng Chen, Tao Wang, Weihao Yu, Yujun Shi, Zi-Hang Jiang,
  Francis~EH Tay, Jiashi Feng, and Shuicheng Yan.
\newblock Tokens-to-token vit: Training vision transformers from scratch on
  imagenet.
\newblock pages 558--567, 2021.

\bibitem{ShiyinZhang2020InteractiveOS}
Shiyin Zhang, Jun~Hao Liew, Yunchao Wei, Shikui Wei, and Yao Zhao.
\newblock Interactive object segmentation with inside-outside guidance.
\newblock In {\em 2020 {IEEE/CVF} Conference on Computer Vision and Pattern
  Recognition, {CVPR} 2020, Seattle, WA, USA, June 13-19, 2020}, pages
  12231--12241. {IEEE}, 2020.

\bibitem{SixiaoZheng2021RethinkingSS}
Sixiao Zheng, Jiachen Lu, Hengshuang Zhao, Xiatian Zhu, Zekun Luo, Yabiao Wang,
  Yanwei Fu, Jianfeng Feng, Tao Xiang, Philip~HS Torr, et~al.
\newblock Rethinking semantic segmentation from a sequence-to-sequence
  perspective with transformers.
\newblock pages 6881--6890, 2021.

\end{thebibliography}
}

\end{document}